\documentclass[conference]{IEEEtran}
\usepackage{times}

\usepackage[numbers]{natbib}
\usepackage{multicol}
\usepackage[bookmarks=true]{hyperref}
\usepackage{graphicx}
\usepackage{amsmath}
\usepackage{amssymb}
\usepackage{booktabs}

\usepackage{soul} 
\usepackage{xcolor} 

\definecolor{dred}{rgb}{0.8,0,0}
\newcommand{\alert}[1]{\textcolor{dred}{\hl{\textbf{#1}}}}
\renewcommand{\alert}[1]{}\renewcommand{\hl}[1]{} 

\begin{document}

\title{Demonstrating a Control Framework \\ for Physical Human-Robot Interaction Toward Industrial Applications}




%
\author{\authorblockN{Bastien Muraccioli\authorrefmark{1}\authorrefmark{3},
Mathieu Celerier\authorrefmark{1}\authorrefmark{3},
Mehdi Benallegue\authorrefmark{1}, and
Gentiane Venture\authorrefmark{1}\authorrefmark{2}}
\authorblockA{\authorrefmark{1}National Institute
of Advanced Industrial Science and Technology (AIST), CNRS-AIST
Joint Robotics Laboratory (JRL),\\ Tsukuba, Ibaraki 305-8560, Japan\\ 
Email: \{firstname.lastname\}@aist.go.jp} 
\authorblockA{\authorrefmark{2}Department of Mechanical Engineering, The University of Tokyo, Tokyo 113-8654, Japan\\Email: venture@g.ecc.u-tokyo.ac.jp}
\authorblockA{\authorrefmark{3}B.Muraccioli and M.Celerier share the first authorship.}}

\maketitle

\begin{abstract}
 Physical Human-Robot Interaction (pHRI) is critical for implementing Industry 5.0, which focuses on human-centric approaches. However, few studies explore the practical alignment of pHRI to industrial-grade performance. This paper introduces a versatile control framework designed to bridge this gap by incorporating the torque-based control modes: compliance control, null-space compliance, and dual compliance, all in static and dynamic scenarios. Thanks to our second-order Quadratic Programming (QP) formulation, strict kinematic and collision constraints are integrated into the system as safety features, and a weighted hierarchy guarantees singularity-robust task tracking performance. The framework is implemented on a Kinova Gen3 collaborative robot (cobot) equipped with a Bota force/torque sensor. A DualShock 4 game controller is attached to the robot's end-effector to demonstrate the framework's capabilities. This setup enables seamless dynamic switching between the modes, and real-time adjustments of parameters, such as transitioning between position and torque control or selecting a more robust custom-developed low-level torque controller over the default one. Built on the open-source robotic control software \texttt{mc\_rtc}, our framework ensures reproducibility for both research and industrial deployment, this framework demonstrates a step toward industrial-grade performance and repeatability, showcasing its potential as a robust pHRI control system for industrial environments.
\end{abstract}

\IEEEpeerreviewmaketitle

\addtocounter{footnote}{-1}
\begin{figure}
  \centering
  \includegraphics[width=1\linewidth]{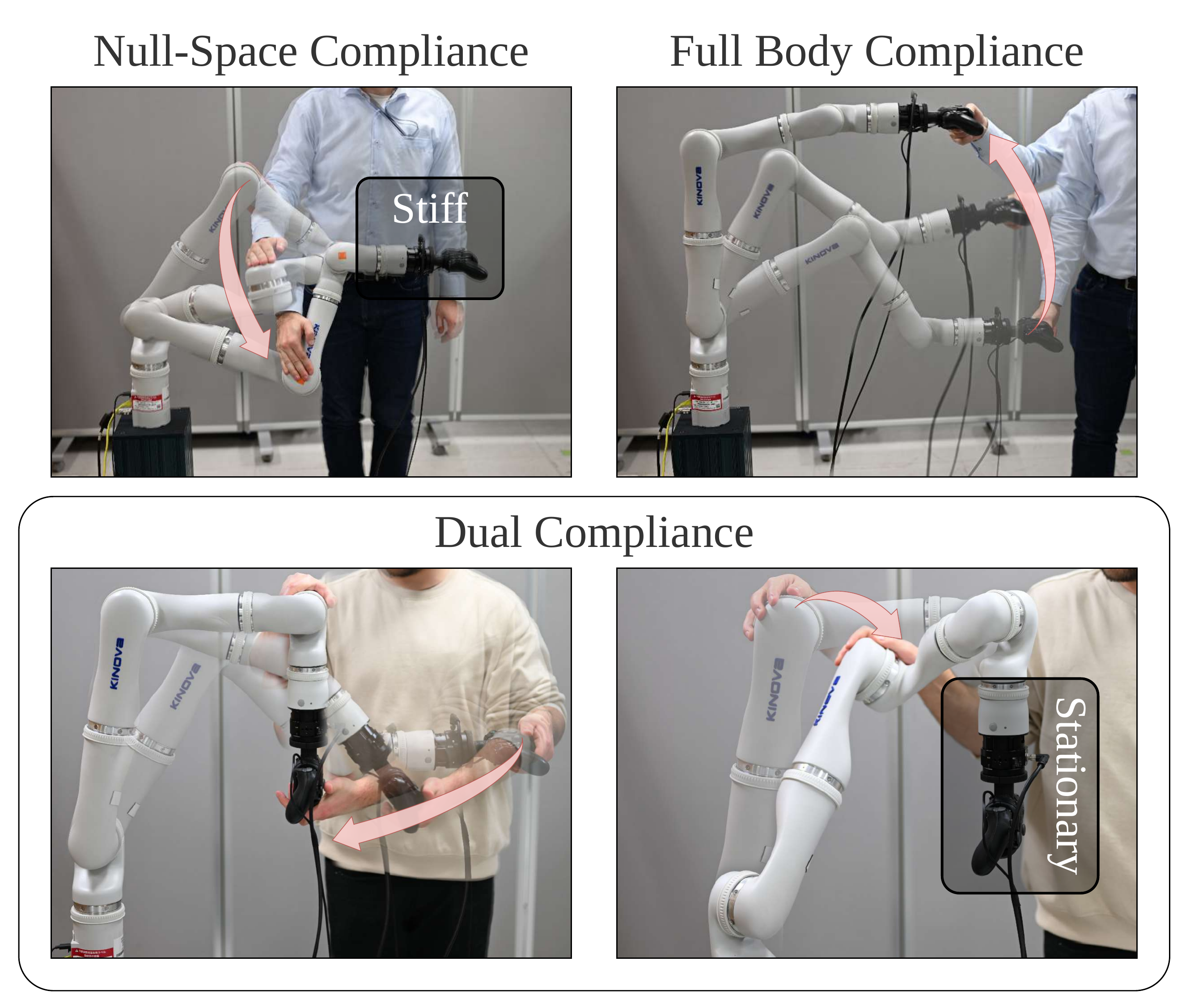} 
  \caption{The different compliant control modes for physical human-robot interaction (pHRI) demonstrated in this paper.
  Top left: Null-space compliance, where the main task (e.g., the pose of the end-effector) is stiff, and compliance is restricted to the null-space of the task. Top right: Full-body compliance, where both the main task and the null-space comply. Bottom: Dual compliance, where the main task becomes compliant only when interacting with the relevant part of the body involved in the kinematic task, and remaining stationary otherwise, while the robot remains compliant in the null-space.}
  \label{fig:modes}
\end{figure}



\section{Introduction}

The emergence of Industry 5.0 emphasizes close physical collaboration between humans and robots~\cite{su11164371}, necessitating robot control policies that can concurrently guarantee safety to protect workers operating alongside robotic systems~\cite{Panagou2023ASR} and capable of generating precise and repeatable motions to align with industry requirements. While significant research has facilitated the integration of robots into industrial environments, achieving safe and reliable pHRI remains challenging~\cite{hrcindustrialsurvey}.

Torque control used in pHRI has proven efficient in addressing safety challenges by enabling compliant behaviors and allowing workers to physically interact with the robot and guide it during operation~\cite{phrihaddadin}. However, this compliance makes the robot more sensitive to environmental factors, which makes the robot's motions less consistent and the tasks less repeatable. By contrast, high-gain position control, widely used in industry, has long fulfilled the need for reliability and precision but it makes the robot stiff and unable to adapt to external forces, posing a potential risk of injury if a worker interferes with the robot's trajectory.
In this work, we use the terms stiffness and compliance in the same broader sense as~\cite{ralcelerier}, where these concepts refer to the robot’s physical response to external forces, encompassing both impedance-based control and effective inertia. These terms are used to qualitatively describe how the robot physically reacts to external forces. A stiff system resists such forces, resulting in strong opposing reactions and minimal movement. In contrast, a compliant system is more yielding, allowing greater motion in response to the same disturbance. 

For pHRI to be successfully integrated into industrial environments, according to~\cite{hrcindustrialsettings}, three primary challenges need to be addressed: \textit{human safety}, \textit{productivity}, and \textit{trust-in-automation}. Human safety requires robust control mechanisms that prevent collisions and mitigate risks in human-robot collaboration, such as implementing compliance control and torque limitations. Productivity depends on designing systems that maintain high precision and efficiency without sacrificing flexibility in adapting to human intervention. Lastly, trust-in-automation demands robots that behave in predictable and task-specific ways that align with the human operators' expectations, fostering confidence in their reliability and usability.

The pHRI framework presented in this demonstration addresses these challenges by incorporating safety features comparable to the reliability and robustness of traditional industrial robotic systems. While \textit{trust-in-automation} is crucial for the broader adoption of pHRI, this framework focuses primarily on the control architecture rather than task planning and Human behavior modeling, which are beyond the scope of this paper. Nevertheless, the compliance and safety features presented here may contribute to fostering trust by enabling reliable interactions~\cite{hrcindustrialsettings}. 
Additionally, the emphasis is placed on achieving industrial-grade performance by presenting several torque-based compliance control modes specifically designed for pHRI in industrial applications.

Our framework implements null-space compliance with high-accuracy task tracking, which ensures that the robot's posture remains flexible while maintaining stiffness and precision at the main task (e.g. end-effector pose), and an implicit compliance control mode, which can be configured for both static and dynamic scenarios. In the dynamic mode, compliance is maintained while the robot is in motion. Furthermore, we introduce a new implementation of the dual compliance mode, as first proposed in~\cite{dualcompliance}, offering greater flexibility for complex collaborative tasks and kinesthetic teaching.

The paper's major contribution is an open-source, industrial-grade pHRI framework that implements several compliant control modes, based on a second-order Quadratic Programming (QP)~\cite{torque-control-QP} formulation to have strict kinematic and self-collision constraints while ensuring singularity-robust task tracking through a weighted hierarchy. In addition, we introduce a novel low-level torque controller for the Kinova Gen3 collaborative robot (cobot), which achieves precision similar to position control while maintaining compliance. \\

Section \ref{sec:relatedworks} discusses the related works and positions the framework in the existing literature. Section \ref{sec:framework} provides a high-level overview of the proposed framework, while Section \ref{sec:hardware} describes the hardware setup. Section \ref{sec:lowlevel} details the design of our new low-level control. Section \ref{sec:implementation} outlines the implementation of the framework, including the compliance modes and a \texttt{mc\_rtc} introduction, followed by Section \ref{sec:evaluation}, which validates the system through experiments. Section \ref{sec:limitations} discusses limitations, and Section \ref{sec:conclusion} suggests future research directions and concludes with a summary of the contributions. This paper is complemented by a website\footnotemark[\value{footnote}] showcasing videos of the framework and includes a link to the GitHub repository of the source code.

\section{Related Works}
\label{sec:relatedworks}

Several pHRI frameworks have been introduced in recent years, with early research primarily focused on collision detection and reaction mechanisms to prevent injuries caused by robots in industrial settings~\cite{4650764}. Over time, more advanced approaches have emerged, integrating risk metrics into robotic control systems to account for injury risks during human-robot interactions~\cite{phrihaddadin}. Additionally, innovations such as precise but heavy compliance achieved through admittance control on industrial robots~\cite{8793657} and the use of soft skins with embedded force sensors~\cite{yan2024soft} have made robots more adaptable to close-contact interactions.

However, despite these advancements, the performance of these systems has often fallen short of industry requirements. Industry-grade performance typically refers to compliance with established performance and safety standards, such as ISO 9283~\cite{ISO9283:1998}, which defines key criteria like pose and path accuracy, static compliance, and stabilization time for industrial robots, and ISO/TS 15066~\cite{ISO15066:2016}, which outlines safety thresholds for collaborative robotics. Furthermore, recent work such as~\cite{treeofrobots} proposes a classification framework for robots based on fitness metrics (e.g., precision, force capabilities) to evaluate their suitability for specific processes through decomposition into sub-processes or tasks, each of which depends on a subset of identified fitness metrics.

This highlights the need to shift the approach toward performance-oriented solutions. Instead of viewing safety as a limiting factor, the framework presented in this paper positions safety as a constraint within performance optimization, as suggested in~\cite{VILLANI2018248}, ensuring both high safety standards and enhanced performance in an industrial pHRI context. By demonstrating in Section~\ref{sec:evaluation} that our torque control strategy achieves motion‑tracking precision on par with the Kinova Gen3’s default position controller, we consider this work a step toward industrial application. 


\section{Features Overview}
\label{sec:framework}
\begin{figure*}
    \centering
    \includegraphics[width=0.75\linewidth]{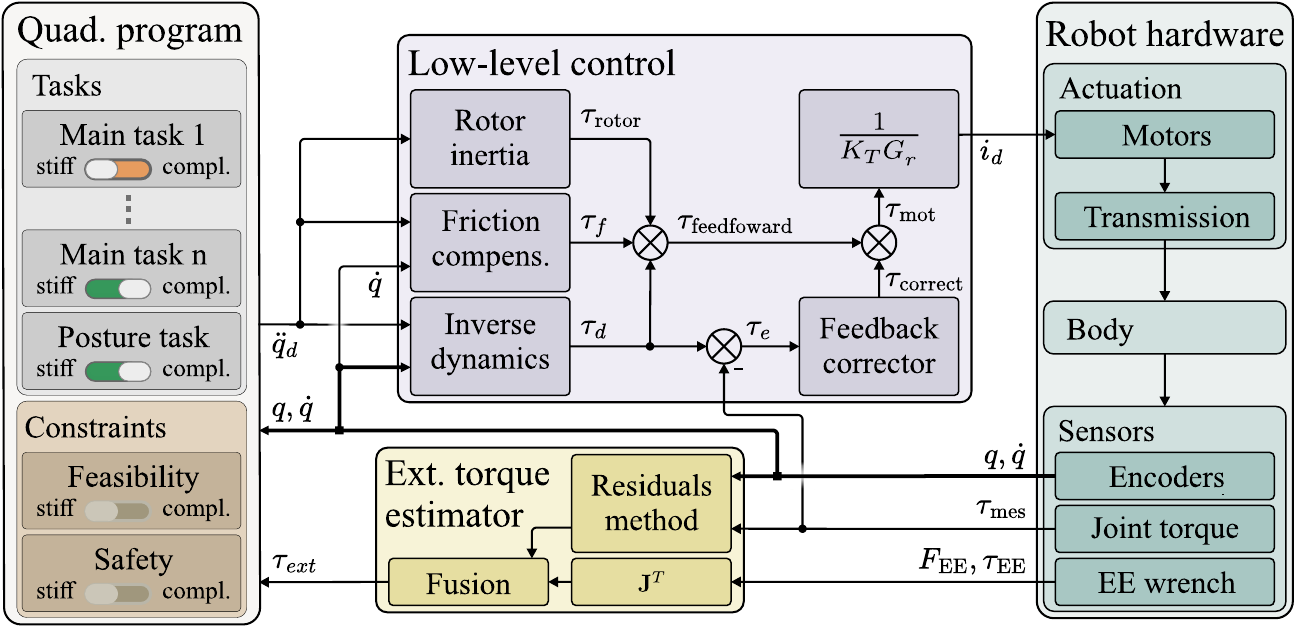}
    \caption{Overview of the proposed control framework. The light-gray block represents the tasks and constraints in the QP. The light-purple block denotes the low-level torque control, ensuring desired joint acceleration and torque tracking by controlling motor currents. The light-green block corresponds to the robot hardware, including actuators and sensors. The yellow block represents the computation of external forces, which are fed into the QP.}
    \label{fig:global-system}
\end{figure*}

Our framework can seamlessly switch between position control to illustrate traditional industrial setup and torque control, essential for pHRI~\cite{farajtabar2024pathcontactbasedphysicalhumanrobot}. In both control modes, the robot can operate in either a static manner, where it maintains a fixed position, or a dynamic manner, where it performs tracking tasks, for demonstration, we implemented a Cartesian task that generates a sinusoidal trajectory.
While effective for precision tasks, position control renders the robot stiff and unable to respond to external forces. This rigidity poses significant safety risks in collaborative environments. 
Our control framework overcomes this limitation by offering different compliant behaviors while incorporating safety constraints that enable safe operation in close proximity to humans in industrial environments, and while maintaining the same precision as position control.

\subsection{Control Modes}

As illustrated in Fig.\ref{fig:modes}, the framework incorporates three distinct compliant control modes, each tailored for specific collaborative scenarios:
\begin{enumerate}
\item \textbf{Null-Space Compliance}
In this mode, compliance is limited to the null-space of the robot. The null-space refers to the set of joint configurations or motions that do not affect the main task (e.g., end-effector position and orientation in Cartesian space). By making the null-space compliant, the robot can dynamically adapt its posture to avoid obstacles or adjust to human intervention without disturbing the main task. This capability is instrumental in cooperative manipulation tasks, where the robot must ensure safety while maintaining task accuracy. \\

\item \textbf{Full-Body Compliance}
This mode extends compliance to the entire robot, including the main task. Unlike null-space compliance, where only the null-space can adapt, full-body compliance ensures that the entire robot responds to external forces. This mode is suitable for scenarios where human operators need the ability to physically interrupt or modify a task, either intentionally or unintentionally, without compromising safety.
The distinction between null-space compliance and full-body compliance provides flexibility in tailoring the robot's behavior to the specific requirements of the task. \\

\item \textbf{Dual Compliance}
The third mode, introduced in~\cite{dualcompliance}, combines the benefits of both null-space and full-body compliance. In dual compliance mode, the body of the main task remains easy to move by the user when touched. However, if the user interacts with the robot on another body, only the posture will adapt, leaving the main task tracking unaffected. This decoupling of null-space and main task compliance makes the dual compliance mode particularly convenient for interactive robot positioning such as kinesthetic teaching. By maintaining a clear distinction between the robot's posture (null-space) and the task being performed (end-effector), this mode simplifies adjustments to the complete robot’s configuration.
\end{enumerate}

\subsection{Safety constraints}
The proposed framework incorporates safety constraints within the second-order QP formulation, ensuring compliance with human-robot collaboration standards. This QP formulation optimally computes joint accelerations, which are then used in the torque control process. The primary safety objectives include enforcing torque limits to adhere to industrial safety regulations and maintaining joint velocity and position limits to prevent self-collision and excessive movement speeds.

To manage these constraints, a second-order velocity damper is implemented as an extension of the first-order approach. This method dynamically adjusts the robot's motion based on distance constraints, ensuring a smooth transition when approaching safety limits. The adaptation to a second-order formulation allows tuning of the system's damping properties. By ensuring an overdamped response, oscillations are prevented, and smooth transitions are maintained.

Additionally, velocity limits are enforced by constraining the acceleration such that the robot remains within predefined velocity boundaries. 

Finally, self-collision avoidance is implemented in Cartesian space, following the same principles as the joint velocity damper.

One important property of these safety features is that they are "stiff", in the sense that they don't only ensure that their own dynamics leads to constraint violation, but also \emph{negate} external forces and torque that push toward these constraints.

The implementation details of the control modes and safety features are detailed in Section \ref{sec:implementation}.


\section{Hardware Setup}
\label{sec:hardware}
The demonstrated framework can run on various manipulator robot arms and utilize different Force/Torque sensors. Here, we present the hardware setup of the proposed demonstration. We implemented our framework on a Kinova Gen3, a 7 Degrees of Freedom (DoF) redundant collaborative manipulator robot arm equipped with joint torque sensors. Kinova's \emph{Kortex} Application Programming Interface (API) allows joint-level Position, Velocity, Torque, and Current control of the robot at 1 kHz in what Kinova refers to as "low-level servoing". We equipped the robot's end-effector with a BOTA SensONE Force/Torque (F/T) sensor, which exhibits very low drift for precise and instantaneous measurement of the forces applied at the end-effector. The F/T sensor communicates via the EtherCAT protocol, and we connected a Next Unit of Computing (NUC) as an EtherCAT master; the data from the force sensor are then published through ROS (Robot Operating System) topics.

The NUC and the robot are connected by Ethernet cables through a switch to a laptop (control PC) running our framework.
For intuitive and straightforward interaction with the different control modes, we used a DualShock 4 (DS4) controller mounted via a custom 3D-printed support and connected it via USB to the control PC. The DS4 controller serves as an optional user interface, allowing the operator to switch between modes in real time. The DualShock 4 is not required for the framework to function; it can be extended to various end-effector tools, such as industrial grippers, screwdrivers, or welding devices.
While this controller may not be perfectly modelled, the system still enables precise control during the demonstration. This further illustrates the robustness of our approach to real-world inconsistencies and hardware imperfections. \\

Although the presented demonstration can run on any manipulator robot arm, this framework strongly relies on torque control; therefore, the targeted robot must support joint torque control. It is also essential that the torque control provides good response and accurate tracking of the input reference joint torque. While Kinova’s default torque control schemes do not offer satisfactory tracking of the reference joint torque, our framework can be generalized to other robots that provide similar torque control capabilities. 

Furthermore, even if a robot's current or direct torque control is not optimal, these control mechanisms can be enhanced like that presented in the next section of this paper, thereby enabling the use of the demonstrated framework. To address this limitation, our first contribution is a novel, robust low-level torque controller that leverages the robot's current control to achieve significantly improved tracking performance, thereby ensuring accurate and reliable torque control under various operating conditions.

\section{Low-level Control}
\label{sec:lowlevel}

Kinova Gen3 offers two low-level torque control modes, respectively, called "Torque control" and "Torque control (High-Velocity)". The former introduces negative feedback on the joint velocity, resulting in an over-damped robot and a loss of compliance when reaching higher velocities, while the latter presents poor torque tracking. Kinova's "High-Velocity" torque control is defined as the torque error filter by a 3rd order z-transfer function fed to the current control. We assume this transfer function was built by Kinova to serve a purpose unreleased, therefore, our proposed low-level torque control builds around this transfer function to improve torque tracking. \\ 

Fig.\ref{fig:global-system} shows the block diagram of the proposed low-level torque control which revolves around 5 components, joint torque-to-current conversion, the inverse dynamics, a \emph{friction compensation} model, the rotor inertia, and as a feedback corrector, the Kinova's transfer function and a leaky integrator. 

This control can be divided into two parts, the feedforward component, which sends the desired torque with the rotor inertia compensation and friction compensation, and the feedback component, which processes the torque error and sends it to the Kinova's transfer function and the leaky Integrator. Then, the joint torque computed by both components is converted into motor current using the torque-to-current conversion factor, $\frac{1}{K_T G_r}$, where $K_T$ is the motor torque constant and $G_r$ is the gear ratio. \\

Conventional acceleration-based torque control primarily relies on inverse dynamics. However, since our implementation builds upon current control, we also account for rotor inertia to produce appropriate torques. The effect of rotor inertia is especially important in computing the desired torque for small joints usually at the end of the kinematics chain, as rotor inertia represents a very important part of the whole inertia to be driven.

To improve the torque tracking, we added a friction model that precompensates friction between the motor and torque sensor, usually in the harmonic drive. The following model is employed:
\begin{equation}
    \tau_f =
    \begin{cases}
        \tau_c\; \text{sign}(\dot{q}) + \tau_v\; \dot{q} & \text{if} \mid\dot{q}\mid  \geq  \dot{q}_{th} \\
        \tau_s\; \text{sign}(\ddot{q}_d) & \text{if}  \mid\dot{q}\mid < \dot{q}_{th} \text{ and } \mid\ddot{q}_d\mid \geq \ddot{q}_{th}\\
        0 & \text{if} \mid\dot{q}\mid < \dot{q}_{th}   \text{ and } \mid\ddot{q}_d\mid < \ddot{q}_{th}
    \end{cases}
\end{equation}

When the velocity falls below a specific threshold, the joint is considered to be in a quasi-static state, implying that the direction of friction depends solely on the desired acceleration. Under such conditions, if the desired acceleration provided by the high-level controller exceeds a second threshold, the friction is compensated in the same direction as that acceleration. Since this acceleration explicitly incorporates external forces, the method enables the robot to precompensate for static friction, even when it is being manipulated by a human. This approach is particularly interesting compared to traditional compliance schemes. In conventional compliance control, when an external force is applied, a position error arises between the current and target positions. The target is then adjusted based on the force measurement, causing the commanded acceleration to oppose the applied force due to the induced position error. \\

Although, the improved feed-forward helps, due to modeling errors in the identified torques or unmodelled friction, this alone is insufficient. To further improve the torque tracking, we add a feedback loop on joint torque measurement.

 The first part of the feedback loop makes use of the same function used in Kinova's default "High-Velocity" torque control. We believe that this transfer function serves a purpose that hasn't been disclosed by Kinova, and therefore chose to include it in the feedback loop. This transfer function is relatively poorly documented, however, some information provided by Kinova can be found. We know that the transfer functions used by Kinova for their low level are expressed in the Z-domain, and have a general form of 5 poles and 5 zeros:
\begin{equation}
    H_{tf}(z) = \frac{\sum_{i}^{5} b_{zi}z^{-i}}{\sum_i^5 a_{zi}z^{-i}}
\end{equation} 

The actual coefficients $a_{z,i}$ and $b_{z,i}$ of the transfer function we used, were obtained through Kinova's \emph{Kortex} API for which $a_{z\{3,4,5\}} = 0$ and $b_{z\{3,4,5\}} = 0$ which results in a 3rd order filter with a quasi-static gain of the following form:
\begin{equation}
    H_{tf}(z) = \frac{0.020175 - 0.036975z^{-1} + 0.016917z^{-2}}{1 - 1.975063z^{-1} + 0.97518z^{-2}}
\end{equation}

We implemented this transfer function to filter torque error with the hypothesis that it was built to account for the joint torque sensor's response. \\

Despite closing the loop on torque sensor measurements using this transfer function, it remains insufficient in situations that require precise tracking of the commanded torque to achieve precise kinematic control of the robot. To that extent, we add a leaky integrator of the following form:
\begin{equation}
    H_{leak}(s) = \frac{K_I}{s+\theta}
\end{equation}

The leaky integrator takes the torque error as input with $\theta$, the time constant and $K_I$, the gain, this is then summed with the transfer function. This leaky integrator's parameters are set such that the integration is relatively slow to help with the final convergence of the system, such as in situations where friction prevents precise positioning.


\section{Framework Implementation}
\label{sec:implementation}


\subsection{Compliance Tasks}
\label{subsec:compliance}

In our work, compliance tasks are formulated to allow for controlled compliance at both the end-effector level and within the null-space of the robot's motion. This is achieved through a quadratic programming (QP) formulation that explicitly incorporates compliance into the optimization problem. By doing so, we can independently adjust the stiffness or compliance of specific tasks based on their requirements.

To enable compliance, as described in~\cite{ralcelerier}, we modify the conventional QP formulation by introducing a compliance parameter $\gamma_k$ for each task. This parameter, which lies in the range $[0, 1]$, governs the degree of compliance for a given task. The result of the QP corresponds to the desired acceleration $\ddot{\mathbf{q}}_d$. The modified QP is expressed as:
\begin{align}\label{eq:compliantQP}
\min_{\ddot{\mathbf{q}}^r} && (\scriptstyle{\sum_k}w_k&\Vert\ddot{\mathbf{e}}^r_k-\ddot{\mathbf{e}}^*_k-\gamma_k\mathbf{J}\hat{\ddot{\mathbf{q}}}^e\Vert^2+w_0\Vert\ddot{\mathbf{q}}^r-\ddot{\mathbf{q}}^*-\gamma_0\hat{\ddot{\mathbf{q}}}^e\Vert^2) \nonumber\\
\text{s.t.} && \mathbf{A}_c\ddot{\mathbf{q}}^r&\leq\mathbf{b}_c \nonumber
\end{align}

where $\mathbf{J}_k$ is the Jacobian matrix of the $k$-th task; $\hat{\ddot{\mathbf{q}}}^e = \mathbf{M}^{-1}\hat{\boldsymbol{\tau}}^e$ represents the estimated effect of external forces on joint accelerations; $\ddot{\mathbf{q}}^r$ is the decision variable minimized in the QP; $\ddot{\mathbf{q}}^*$ represents the posture task's reference in accelerations, given by the controller; The pair $\ddot{\mathbf{e}}^r_k$ and $\ddot{\mathbf{e}}^*_k$ are the optimization variable and the reference of the $k$-th task given by the controller; $w_k$ and $w_0$ are respectively the weights of the $k$-th task and the posture task; $\gamma_k$ is the compliance parameter for the $k$-th task; The term $\hat{\boldsymbol{\tau}}^e$ corresponds to the estimated external torques.

The optimization framework allows for explicit compliance tuning by adjusting $\gamma_k$ for each task. For example, setting $\gamma_k=0$ results in a stiff behavior for the operational space tasks, while $\gamma_k=1$ makes them fully compliant. Similarly, the null-space compliance can be controlled by setting the parameter $\gamma_0$ for the null-space task, which is treated as an additional optimization objective with the Jacobian set to identity.

This approach ensures that external forces are effectively incorporated into the control framework. Tasks can exhibit compliance in the presence of external disturbances while retaining their primary objectives. For instance, when $\gamma_k=0$ and $\gamma_0=1$, only the null-space behavior is compliant, allowing for disturbance rejection in the operational space. Conversely, setting $\gamma_k=1$ introduces compliance to the operational tasks, enabling impedance-like behavior.

By leveraging this formulation, we provide a flexible method for balancing stiffness and compliance across multiple tasks. This flexibility is particularly advantageous in scenarios requiring adaptive behavior, such as human-robot interaction or environments with uncertain dynamics. The theoretical framework also ensures that safety constraints, represented by $\mathbf{A}_c \ddot{\mathbf{q}}^r \leq \mathbf{b}_c$ are respected throughout the optimization process. \\

Concerning dual compliance, its introduction in~\cite{dualcompliance}, defined it as a combination of null-space compliance with full-body compliance achieved through admittance control. When the measured force at the end-effector remains below a predefined threshold, the reference Cartesian position of the end-effector is maintained stationary and the null-space is still compliant. Once the measured force exceeds the threshold, full-body compliance control gets activated, allowing the end-effector’s position to be freely modified. Upon release, the new position becomes the updated reference, and the system reverts to null-space compliance only.

In our implementation of the dual compliance, instead of relying on an admittance task as in~\cite{dualcompliance}, we utilize our QP-based compliant end-effector task with zero stiffness and relatively low damping for stability. This setting allows the end-effector to move with minimal force (determined relative to the damping parameter) compared to conventional admittance control. This approach exploits direct compliance, which is intrinsic to the torque control. The new implementation of dual compliance benefits from a more lightweight feel at the end-effector, while keeping precision thanks to our low-level torque control.

The force threshold in our framework is measured using the Bota F/T sensor mounted at the end-effector. To handle transitions between null-space compliance and full-body compliance, we implemented a Schmitt trigger mechanism with two distinct thresholds. When the measured force exceeds the high threshold, the system enters the dual compliance mode, allowing the end-effector to move under applied force. Once the measured force drops below the low threshold, the system reverts to null-space compliance only. This hysteresis-based approach ensures smooth transitions and avoids unnecessary mode switching caused by minor force fluctuations near the threshold.

\subsection{Safety Features}
The proposed framework includes safety constraints using our QP formulation. These safety constraints serve two critical purposes: first, they bound torque limits, to comply with established safety standards for human-robot collaboration~\cite{hrcindustrialsurvey}; second, they ensure that the robot operates within velocity and joint position limits. 

To address these constraints, we implement a second-order velocity damper, an extension of the first-order method introduced by Faverjon and Tournassoud~\cite{velocitydamper} which is defined as:
\begin{equation}
\dot{e}_d = -\xi \cdot \frac{(e - d_s)}{(d_i - d_s)}, \label{eq:first-order-vel-damp}
\end{equation}
where $\dot{e}_d$ represents the velocity limit, $e$ is the current distance, $d_s$ is the safety distance (the minimum allowable distance), $d_i$ is the influence distance (the threshold where the constraint is added to the QP), and $\xi$ is a positive coefficient controlling the convergence speed.

Since our QP formulation computes the joint accelerations $\ddot{\mathbf{q}}_d$, we need to express the velocity damper in terms of the QP decision variable $\ddot{\mathbf{q}}^r$. Therefore, we implemented the second-order velocity damper constraint. This constraint limits the acceleration to a value $\ddot{e}_p$ which is expressed in terms of the error between the current velocity $\dot{e}$ and the desired velocity $\dot{e}_d$ from \eqref{eq:first-order-vel-damp}, with a damping gain $\lambda$, as follows:
\begin{equation}
\ddot{e}_p = -\lambda(\dot{e} - \dot{e}_d) \label{eq:sec-order-vel-damp}
\end{equation}

This formulation was already presented in~\cite{vaillant2014vertical} and was tuned to guarantee immediate velocity convergence by setting $\lambda=\dfrac{1}{\delta_T}$ where $\delta_T$ is the control sampling time. This setting is acceptable for position control, which usually has an ideal open-loop joint acceleration integration. On the other hand, closed-loop inverse dynamics control requires to be robust to noises, joint tracking errors, and model discrepancies. In this case, such a high value for $\lambda$ would cause chattering and would amplify high-frequency noises.

To determine $\lambda$ in a better way, The position constraint has been rewritten in the form of a second-order system by combining \eqref{eq:first-order-vel-damp} and \eqref{eq:sec-order-vel-damp}. We tune the gains of of this second order system such that the amortization margin $M$ is greater than $1$, which ensures the system is overdamped in order to robustly guarantee smoothness and prevent oscillations.

Accordingly, the damping gain $\lambda$ is then determined by:
\begin{equation}
\lambda = \frac{4M^2 \cdot \xi}{(d_i - d_s)},
\end{equation}

Then, substituting again in \eqref{eq:sec-order-vel-damp}, the position constraint $\ddot{e}_p$ becomes:
\begin{equation}
\ddot{e}_p = - \frac{\lambda^2}{4M^2} \cdot (e - d_s) - \lambda \cdot \dot{e},
\end{equation}

In a comparable way, we can also enforce the robot's velocity limits $\dot{e}_{\text{lim}}$ by bounding the acceleration with a limit $\ddot{e}_v $. The expression is simpler than for the velocity damper since we replace the reference $\dot{e}_d$ in \eqref{eq:sec-order-vel-damp} with $\dot{e}_{\text{lim}}$, giving
\begin{equation}
\ddot{e}_v = -\lambda \cdot (\dot{e} - \dot{e}_{\text{lim}}),
\end{equation}

At last, both constraints are applied to the QP in the form of bounding values of the acceleration command $\ddot{q}_d$:
\begin{equation}
\ddot{q}_d \leq \min(\ddot{e}_p, \ddot{e}_v) 
\end{equation}
in the case of an upper limit and
\begin{equation}
\ddot{q}_d \geq \max(\ddot{e}_p, \ddot{e}_v)
\end{equation}
in the case of a lower limit. 

One important property of the safety constraint is that it is stiff, in the sense that since the predicted acceleration $\ddot{q}_d$ considers the effect of the external force, it would incorporate its consequences on the alignment with the safety constraints. Therefore, any force that would put these constraints at stake would be negated as soon as it is detected unless the robot reaches its torque limits.

In addition, we implement (self-)collision avoidance by continuously monitoring the minimum distance between pairs of robot bodies and generating damping behaviors as this distance approaches a predefined safety threshold. This is based on the velocity damper implementation of~\cite{vaillant2016multi} and~\cite{qp-constraint}, where the variable $e$ describes the distance between the closest points on each body pair.
Further details of these safety guarantees are presented in~\cite{ralcelerier}

\subsection{The control framework \texttt{mc\_rtc}}
\label{subsec:mcrtc}


\texttt{mc\_rtc}~\cite{mcrtcfsm} is an open-source software framework designed for the development, simulation~\cite{singh2022mcmujocosimulatingarticulatedrobots}, and deployment of model-based control strategies for robots. It features a Second-Order QP solver, which enforces strict constraints while enabling task prioritization through a weighted hierarchy. This ensures robust task tracking, even in singularity-prone scenarios, making it an ideal foundation for implementing compliant control strategies in pHRI. \\

\texttt{mc\_rtc} provides a wide array of pre-defined tasks, including end-effector positioning, posture control, visual servoing, force control, admittance tasks, and more. These tasks can be executed in open-loop control or, as in our case, in closed-loop configurations that integrate real-time sensor feedback. \texttt{mc\_rtc}'s flexibility allows us to integrate the compliance algorithms described in this paper, into both posture and end-effector tasks. Moreover, the same controller can be deployed across different robots and simulation environments without rewriting the controller. New platforms can be integrated using their Unified Robot Description Format (URDF), enabling rapid development and testing without requiring significant modifications to the controller.
Since \texttt{mc\_rtc} is designed to be robot-agnostic, the proposed framework can be easily adapted to a wide range of robotic platforms with kinematic redundancy, such as 6- or 7-DoF arms. These features could help the industry to efficiently adapt our framework to their needs, test it, and validate it on their robots and hardware in the factory setting.
For our pHRI framework, \texttt{mc\_rtc} allowed us to control the Kinova Gen 3 at 1kHz, ROS was used for communication and interface, while MuJoCo served as a simulation environment during the initial development phase. \\

Additionally, we used \texttt{mc\_rtc} to implement a Finite State Machine (FSM)~\cite{mcrtcfsm} controllable via a DualShock 4 game controller, enabling real-time switching between different compliance modes (null-space, full-body, dual compliance), low-level control modes (position control, Kinova's default torque control, and our new torque control), and operational scenarios (static or dynamic). We designed an FSM to demonstrate the capabilities of our framework and its contributions. However, \texttt{mc\_rtc}'s FSM is versatile enough to be applied in diverse industrial applications, such as assembly, inspection, or material handling, where flexible control strategies are required.

\texttt{mc\_rtc} also provides extensive monitoring capabilities, including live plotting and an advanced logging system. These features enable continuous monitoring of robotic metrics, which is particularly valuable for industrial applications, where reliable performance tracking is critical.

\section{Evaluation}
\label{sec:evaluation}


\subsection{Friction parameters identification}
\begin{figure}[t]
    \centering
    \includegraphics[width=\linewidth]{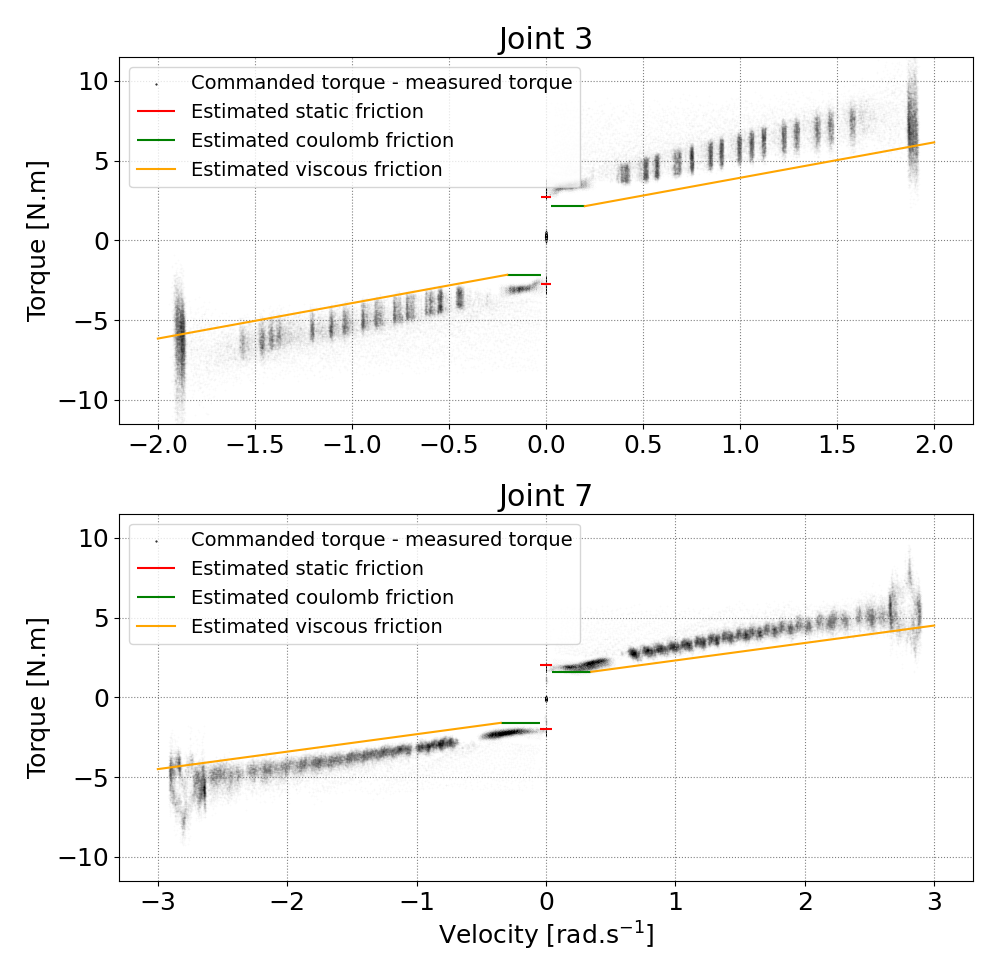}
        \caption{Friction identification data for Kinova Gen3. In shaded grey the actual data, in green the Coulomb friction part, in yellow the viscous friction part and we added a potential estimation of the static friction.}
    \label{fig:friction}
\end{figure}
We identified the friction model parameters from experimental data of torque error between motor torque obtained from motor current, joint torque, and joint velocity measurements. These experimental data were executed while the robot was positioned a zero-torque stance, extending the robot vertically such that no joints get affected by the gravity. We applied constant current on each continuous joint one by one and repeated for multiple current values to cover the whole range of joint velocities. Fig.\ref{fig:friction} shows two examples of friction plots of the torque error plotted against the velocity based on the obtained experimental data on which we overlaid the estimated model of Coulomb and viscous friction. We also identified the static friction, however, we considered the relative difference between the static and the Coulomb friction to be small enough to model them with the same value. From these experimental data, we identified the two following models for the first 4 joints and the last 3 joints referred to by Kinova as "\emph{large}" and "\emph{small}" actuators respectively:
\begin{equation}\label{eq:friction-model}
    \tau_f =
    \begin{cases}
        2.15\; \text{sign}(\dot{q}) + 2.00\; \dot{q} & \text{for large actuators} \\
        1.60\; \text{sign}(\dot{q}) + 1.36\; \dot{q} & \text{for small actuators}
    \end{cases}
\end{equation}

\subsection{Torque tracking}
\begin{figure}[t]
    \centering
    \includegraphics[width=1\linewidth]{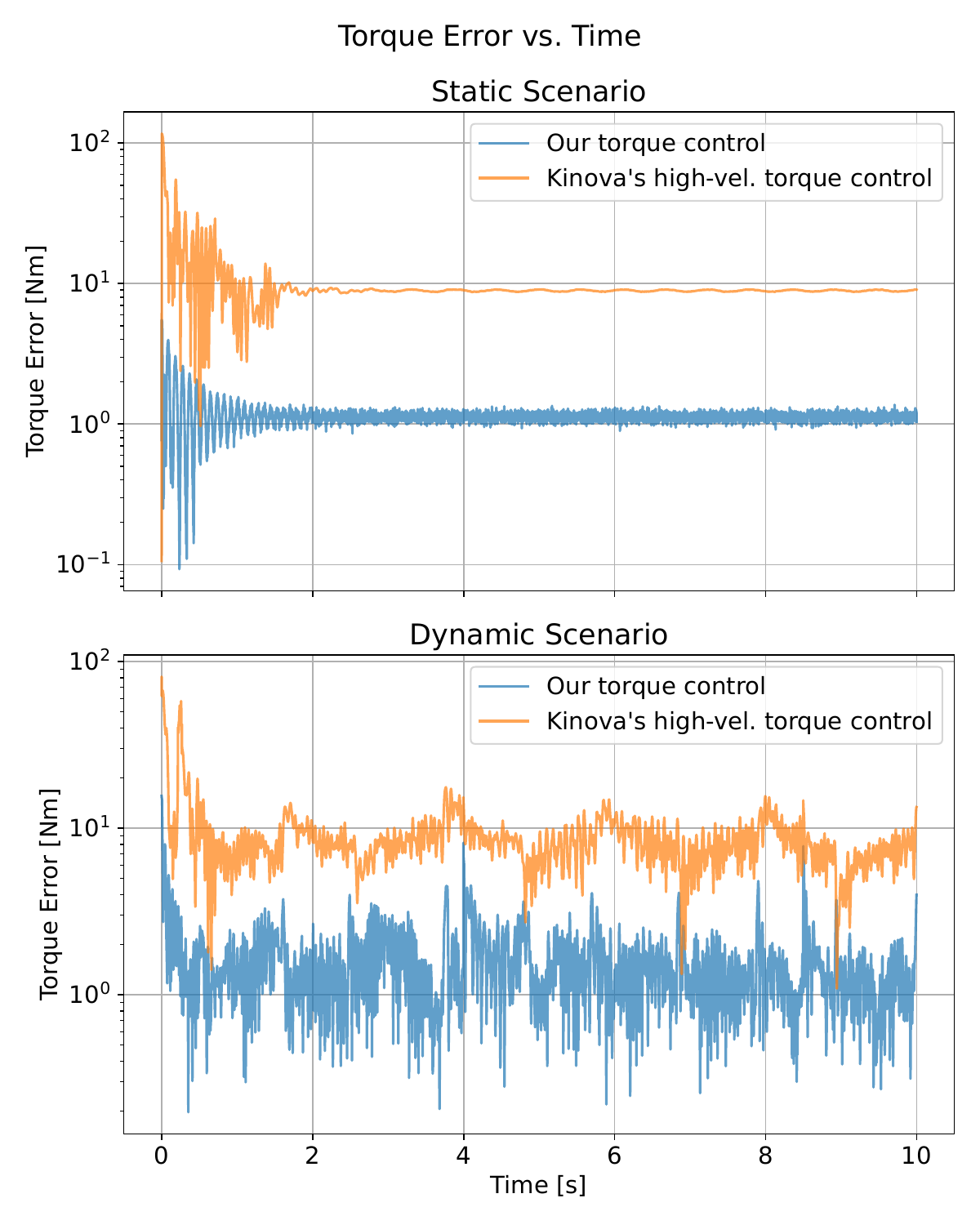}
    \caption{Comparison of the norm of the joint torque tracking error for a static scenario (top) and a dynamic scenario (bottom). Both subplots compare our torque controller (blue) and Kinova’s high-velocity torque controller (orange). The y-axis is presented in a logarithmic scale to highlight differences in the error magnitude.}
    \label{fig:torque-control-results}
\end{figure}
We compared the proposed torque control with Kinova's high-velocity torque control in a static and dynamic scenario. The static scenario consists of partially extending the robot horizontally in position control and then switching to our torque control. We attached a weight at the end-effector such that the joint friction has a small impact on the torque error measurement, we defined the task at the end-effector to be non-compliant such that the weight measured by the F/T sensor is compensated. The dynamic scenario consists of a 6D-Cartesian task moved in a 2-plane doing back-and-forth sinusoidal motion. Fig.\ref{fig:torque-control-results} shows the torque error measurement in both scenarios, for the demonstrated low-level torque control and Kinova's high-velocity torque control, to improve reading of the differences in error magnitude the torque error axis is presented in a logarithmic scale.

\subsection{Comparison with position control}
When considering pHRI, we need a robot that becomes compliant and deviates from its target/trajectory under external forces. However, in some scenarios, it's desirable to keep the main task stiff, which can be a 6D pose tracking or to constrain it only in a specific way (e.g. along a path, on a given Cartesian plane). This critical part of the task needs then to be stiff, while it is still desirable that the remaining degrees of freedom are compliant. The performance of the controller is then measured by the quality of the tracking of that critical task. 

The following results compare our torque control as an example of a robust low-level implementation design, with position control, which remains the standard approach in manufacturing industries.

\subsubsection{Stiffness in static configurations}
\begin{figure}[t]
    \centering
    \includegraphics[width=1\linewidth]{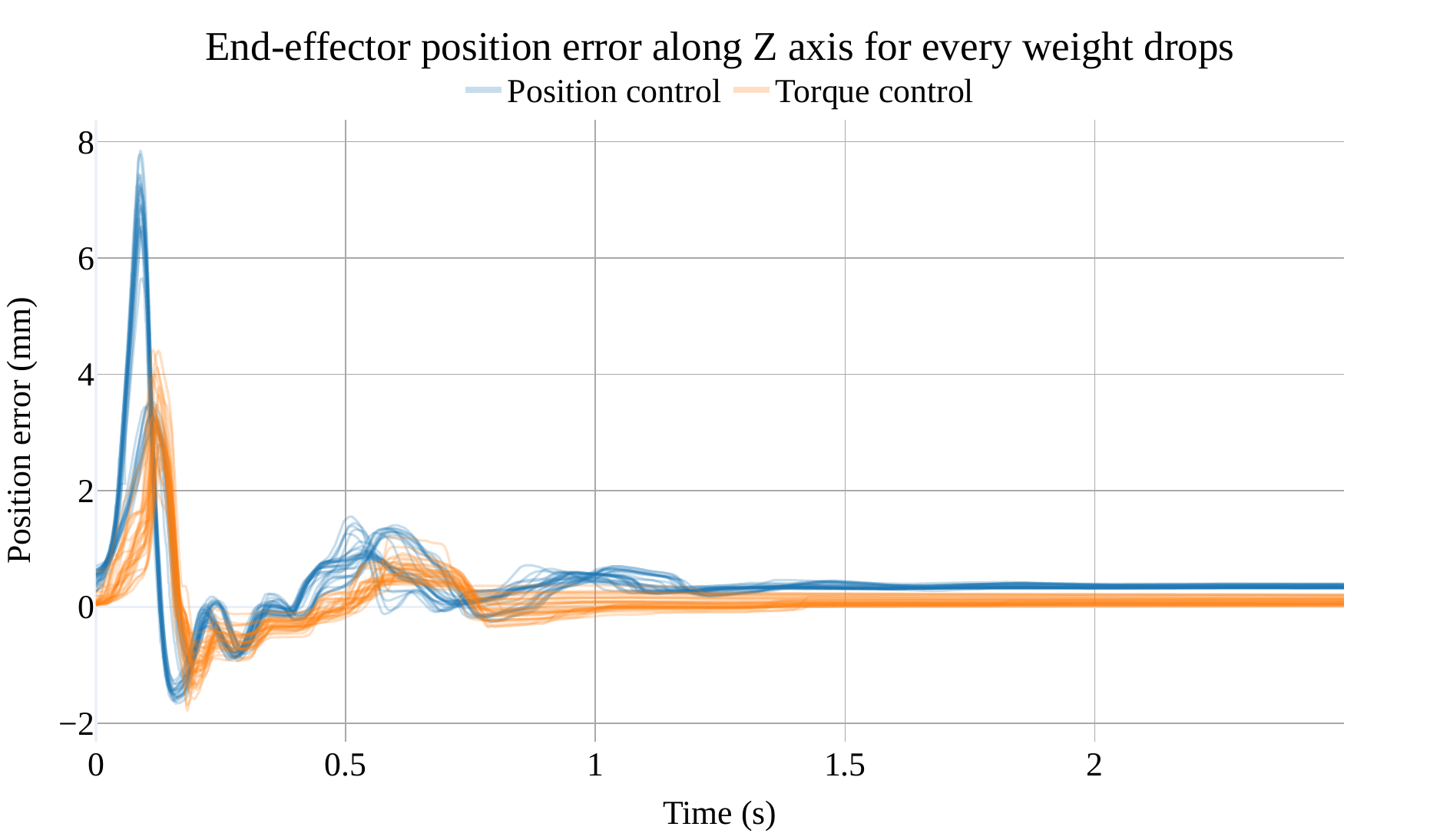}
    \caption{Position error over drop segments for torque and position control strategies. The top subplot represents the position error during torque control, while the bottom subplot represents the position error during position control. Each line corresponds to an individual drop segment. The position error is measured in millimeters (mm) and is plotted as a function of time, with each segment synchronized from the F/T sensor's force measurement. The data illustrate the differences in error behavior between the two control strategies, highlighting the capability of torque control to perform at least as well as position control.}
    \label{fig:static-torque-position-comparison}
\end{figure}
We first compare position control and torque control in a static scenario. We extended the robot's arm horizontally, and we required the robot to keep the 6D pose of its end-effector constant. The two controllers were the manufacturer's high-gain position control and our torque control with stiff end-effector pose task and null-space compliance. To test the capacity of both targets to track the task precisely, we introduced disturbances by attaching a 1.25kg weight at the end-effector of the robot and letting it drop vertically by 25 cm.
To analyze the performance of both control modes, we observed the error between the target position and the real position of the 6D-Cartesian task along the axis impacted by the drop. We repeated 30 weight drops for each control mode and computed for each drop segment the peak error, the time taken for the error to stabilize, and the residual/static error.

\begin{table*}[t]
\centering
\begin{tabular*}{\textwidth}{l@{\extracolsep{\fill}}cccccc}
\toprule
\textbf{Metric} & \textbf{Mean (Torque)} & \textbf{Std (Torque)} & \textbf{Mean (Position)} & \textbf{Std (Position)} & \textbf{p-value} & \textbf{Effect Size} \\
\midrule
Peak & $3.216$ & $0.594$ & $5.632$ & $1.820$ & $1.028e^{-06}$ & $0.736$ \\
Stabilization Time & $0.512$ & $0.226$ & $0.842$ & $0.220$ & $9.513e^{-04}$ & $0.498$ \\
Residual error & $0.102$ & $0.061$ & $0.353$ & $0.025$ & $3.020e^{-11}$ & $1.000$ \\
\bottomrule
\end{tabular*}
\caption{Statistical Analysis Results for static Torque vs. Position Metrics}
\label{tab:stat_analysis}
\end{table*}

The statistical analysis of the torque and position metrics revealed significant differences across all three comparisons (Table \ref{tab:stat_analysis}). For the peak values, the mean torque peak was 3.216 (std = 0.594), while the mean position peak was 5.632 (std = 1.820). The Shapiro-Wilk test indicated that the torque peak data was normally distributed ($p = 0.772$), whereas the position peak data was not ($p \leq  0.001$). Given the non-normality of the position peak data, the Mann-Whitney U test, a non-parametric test, was employed. This test showed a statistically significant difference between the two peaks ($p < 1.028e^{-06}$), with a large effect size (Rank-biserial correlation = 0.736), suggesting that position peaks are substantially higher than torque peaks.

In terms of stabilization times, the mean torque stabilization time was 0.512 seconds (std = 0.226), and the mean position stabilization time was 0.842 seconds (std = 0.220). Both datasets were found to be non-normally distributed ($p < 0.001$), justifying the use of the Mann-Whitney U test. This test indicated a statistically significant difference ($p =  9.513e^{-04}$), with a moderate effect size (Rank-biserial correlation = 0.498), implying that position stabilization times are longer than torque stabilization times.

For the residual values, the mean torque residual was 0.102 (std = 0.061), and the mean position residual was 0.353 (std = 0.025). The torque residual data was normally distributed ($p = 0.070$), while the position residual data was not ($p = 1.078e-03$). Due to the non-normality of the position residual data, the Mann-Whitney U test was used. This test revealed a statistically significant difference ($p < 0.001$), with a very large effect size (Rank-biserial correlation = 1.000), indicating that position residuals are much higher than torque residuals.

Overall, these results highlight substantial differences between the torque and position metrics, with varying degrees of practical significance. The findings suggest that metrics during torque control consistently show lower values compared to their position counterparts, which shows the applicability of torque control, even in industry application that requires high precision/stiffness.

\subsection{Dynamic trajectory tracking}
\begin{figure*}[t]
    \centering
    \includegraphics[width=1\linewidth]{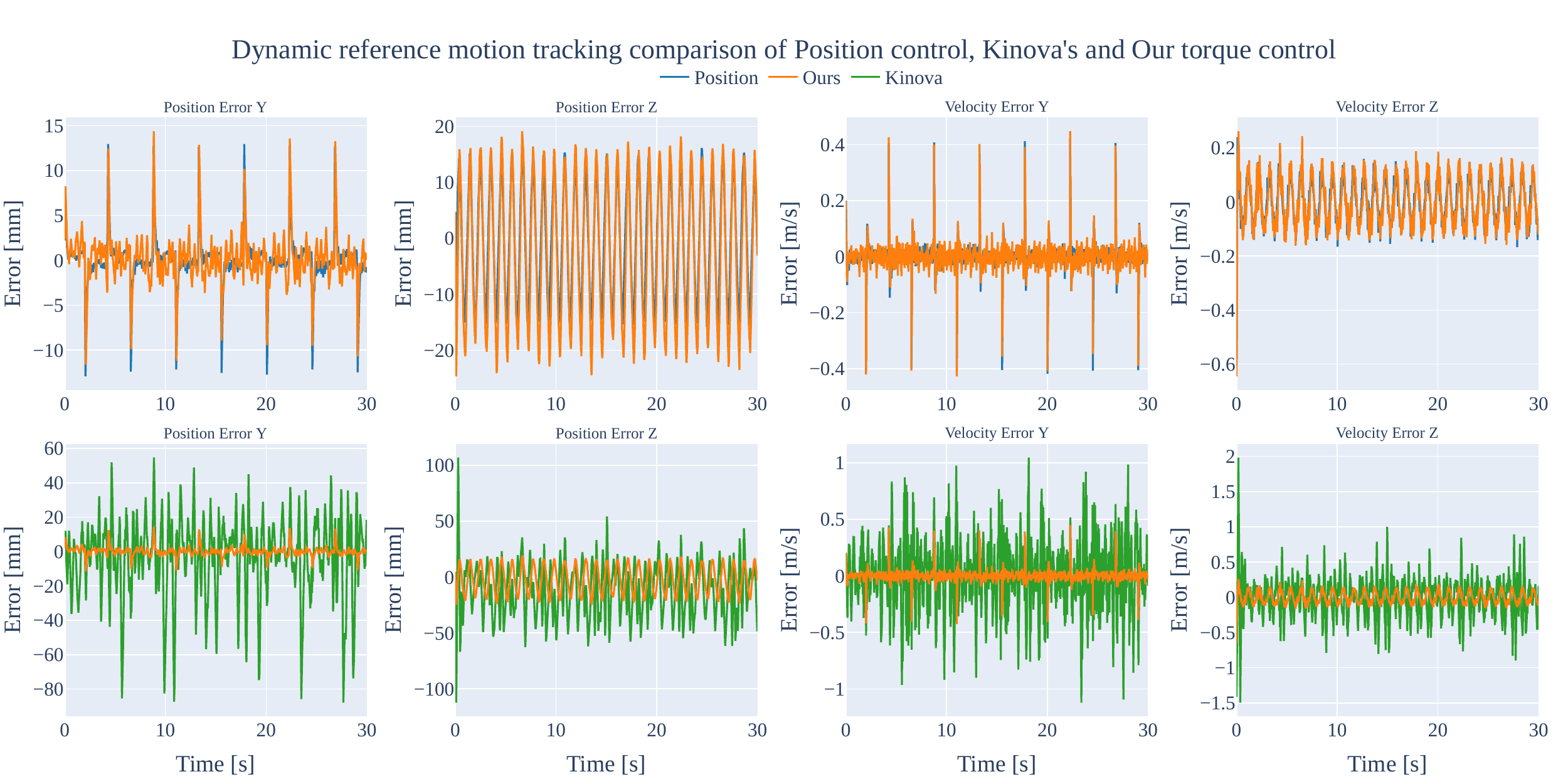}
    \caption{Comparison of dynamic tracking errors between our torque control and position control (top row) and between our torque control and using kinova's torque tracking loop (bottom row). The errors are in terms of position and velocity in both y axis (the horizontal axis) with discontinuous reference velocities and z axis (the vertical axis) with a higher frequency sine function reference.}
    \label{fig:dynamic-comparison}
\end{figure*}
We compared position control and torque control in a dynamic scenario to account for industrial applications where a reference trajectory could be produced by optimization or provided by a planner. We extended the robot arm horizontally and then tracked a 6D-Cartesian trajectory doing back-and-forth sinusoidal motion on the vertical axis and a lower frequency triangular reference on the horizontal axis.
To analyze the performance of both control modes, we observed the error between the reference and real position along the moving axes and did the same for the reference and real velocity by repeating several back-and-forth motions. The results, displayed in Fig.\ref{fig:dynamic-comparison} show that our torque control has comparable tracking precision as position control, while having its null-space compliant. Unsurprisingly, compared to Kinova's torque loop, our torque control shows a much higher precision.

\subsection{Dual compliance}
\begin{figure*}[t]
    \centering
    \includegraphics[width=1\linewidth]{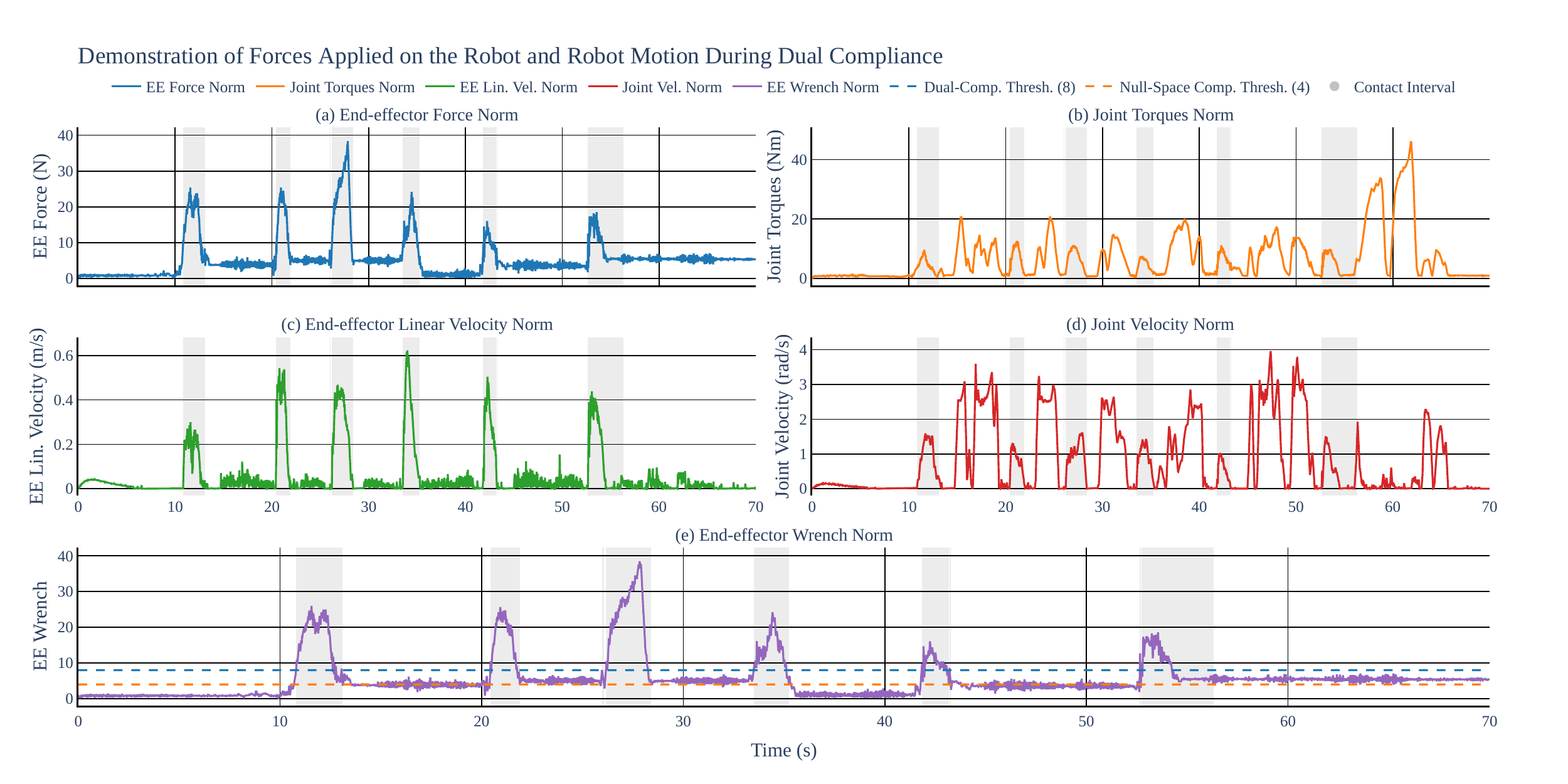}
    \caption{Time-series analysis of robot dynamics during dual compliance operation. (a) Norm of End-Effector (EE) Force, representing the magnitude of forces applied in Newton (N). (b) Norm of Joint Torques, indicating the cumulative torque exerted at each of the seven robot joints in Nm. (c) Norm of EE Linear Velocity, illustrating the speed of the end-effector's linear motion in meters per second (m/s). (d) Norm of Joint Space Velocity, showing the rate of movement within the robot's joint space in radians per second (rad/s). (e) Norm of EE Wrench, encompassing both force and torque magnitudes applied to the end-effector. Shaded regions denote dual compliance interval, identified when the EE Wrench Norm exceeds the initiation threshold of 8 and retracts below the removal threshold of 4. The plots highlight the interplay between the applied forces, joint torques, dual compliance mode switching, and resulting motions during dual compliance control.}
    \label{fig:dual-compliance}
\end{figure*}
In this section, we present the results on the dual compliance strategy. This choice is motivated by its inherent advantages in tasks requiring intuitive and adaptive human-robot interaction, such as teaching a robot a new skill through physical guidance. By allowing the robot to dynamically adjust its stiffness, dual compliance facilitates a more natural exchange of forces, enhancing the robot’s responsiveness to user intent. The following results demonstrate how this approach features low forces for interaction, robot motion, and mode switching dependent on the interaction point. Fig.\ref{fig:dual-compliance} shows the result of an experiment using dual compliance for moving the end-effector of the robot from different positions/orientations and modifying the null-space configuration in between. The mode-switching can be observed in Fig.\ref{fig:dual-compliance}.e), when the end-effector wrench norm becomes greater than the higher (\emph{dashed blue}) threshold, we switch to a compliant end-effector highlighted by the \emph{gray} area. When the wrench norm falls below the lowest (\emph{dashed orange}) threshold, the system switches back to a stiff position and orientation for the end-effector, maintaining only null-space compliance.

\section{Limitations}
\label{sec:limitations}

While the proposed framework demonstrates significant advancements in pHRI toward industry, some limitations remain. One notable challenge is the difficulty in identifying precise friction parameters for the Kinova Gen3 robot. Although the friction model is not yet perfect, our robust dynamic model identification and closed-loop control strategies—enhanced by the torque sensors from the actuators and the F/T sensor feedback—mitigate these imperfections, enabling industry-ready torque control. Nevertheless, further refinement of friction modeling could enhance performance.

Another limitation arises from the lack of transparency in the low-level control architecture of the Kinova Gen3, as the manufacturer provides limited documentation or insight into its internal workings. This constraint necessitated substantial reverse engineering to achieve the results presented in this paper. While these efforts have proven fruitful, they highlight the challenges of developing advanced control strategies on proprietary platforms with limited access to critical system details. A key lesson from this work is the critical importance of incorporating rotor inertia—and, more generally, accurate inertia parameters—into inverse dynamics-based torque control. As discussed in Section~\ref{sec:lowlevel}, since the QP computes a desired acceleration and the corresponding desired torque is obtained from an inverse dynamics model, explicitly integrating rotor inertia into the low-level control significantly improves torque tracking performance when controlling motor torque or current. This underscores the reliance of the proposed approach on precise dynamic model identification, including rotor inertias, friction coefficients, and mass/inertia properties.

One possible way to alleviate this dependency is through adaptive control strategies, which aim to estimate and adjust dynamic parameters online. However, applying such techniques in the context of pHRI remains challenging. During physical interaction, it becomes difficult to distinguish whether deviations in the robot’s behavior stem from external forces applied by a human or from inaccuracies in the dynamic model. This ambiguity complicates the parameter identification process and may lead to incorrect adaptations. Therefore, further research is needed to develop adaptive control methods that are robust to physical contact.

Finally, our framework focuses on control strategies for pHRI but does not yet integrate human-centric evaluation metrics such as user experience, cognitive load, or long-term usability. While the framework improves safety and performance, it has not been formally tested with industrial workers or in real-world collaborative settings. Future studies should incorporate human-in-the-loop evaluations, similar to those in~\cite{CORONADO2022392}, to assess how effectively workers can interact with and control the robot in practical tasks.

Another core pillar of Industry 5.0 is environmental sustainability. While this work primarily focuses on enhancing safety and productivity, future research should address the environmental impact of pHRI systems. Exploring energy-efficient algorithms, sustainable hardware design, and eco-friendly manufacturing processes is crucial for aligning pHRI frameworks with sustainability goals. The methodologies presented in~\cite{BONELLO2024282} provide a valuable foundation for integrating environmental considerations into the design and deployment of robotic systems.

\section{Discussion and Conclusion} 
\label{sec:conclusion}

This paper presented a novel open-source pHRI control framework, designed for industrial environments by addressing the key challenges of safety, compliance, and performance. The contributions of this paper include a new low-level torque control for the Kinova Gen3 robot, incorporating friction identification, more precise torque tracking, and torque control precision similar or superior to position control. This advancement positions our framework as leap toward industrial applications, where performance, precision, and reliability are critical.

Additionally, we introduced a novel dual compliance implementation, which combines null-space and full-body compliance and does not require admittance control. Our framework also leverages the \texttt{mc\_rtc} framework for modularity, scalability, and monitoring capability, underscoring its potential for broad industrial applicability.

Industry 5.0 emphasizes the integration of human-centric values such as personalization, human well-being, and environmental sustainability into industrial processes~\cite{su11164371}. Our proposed framework aligns with these principles by addressing key aspects of physical human-robot interaction (pHRI): ensuring human safety through strict safety constraints, enhancing productivity through industrial-grade performance, and fostering trust in automation through the compliance capabilities of our system. These contributions are critical for the broader adoption of pHRI in industrial settings~\cite{hrcindustrialsettings}. Further works relying on this framework are planned for user-studies on personalization and evaluation of pHRI control laws.

In conclusion, these contributions not only advance the state of torque-based control but also demonstrate its viability for widespread adoption in industrial settings. Future efforts should focus on further refining the framework to align with human-centered design for Industry 5.0. Other aspects involving environmental sustainability and societal impact should also be thoroughly investigated. By addressing these dimensions, advanced torque-based control can serve the next generation of industrial collaborative robotics. These areas represent promising directions for ensuring that pHRI systems meet the broader goals of Industry 5.0.

\section*{Acknowledgments}
This research includes results obtained from the project "Programs for Bridging the Gap between R\&D and the Ideal Society (Society 5.0) and Generating Economic and Social Value (BRIDGE)/Practical Global Research in the AI × Robotics Services", implemented by the Cabinet Office, Government of Japan.


\bibliographystyle{plainnat}
\bibliography{references}

\end{document}